\documentclass[letterpaper, 10pt, conference]{ieeeconf} 

\IEEEoverridecommandlockouts

\usepackage{cite}
\usepackage{amsmath,amssymb,amsfonts}
\usepackage{algorithmic}
\usepackage{graphicx}
\usepackage{textcomp}
\usepackage{xcolor}
\usepackage{algorithm}
\usepackage{dblfloatfix}

\def\BibTeX{{\rm B\kern-.05em{\sc i\kern-.025em b}\kern-.08em
    T\kern-.1667em\lower.7ex\hbox{E}\kern-.125emX}}

\graphicspath{{images/}}

\begin{document}

\title{Target Detection, Tracking and Avoidance System for Low-cost UAVs using AI-Based Approaches}

\author{Vinorth Varatharasan$^{1}$, Alice Shuang Shuang Rao$^{1}$, Eric Toutounji$^{1}$, Ju-Hyeon Hong$^{1}$, Hyo-Sang Shin$^{1}$
\thanks{$^{1}$ School of Aerospace, Transport and Manufacturing, Cranfield University, Cranfield MK43 0AL, UK
(email: \texttt{vinorth.varatharasan@gmail.com}, \texttt{hong.ju.hyeon11@gmail.com})}
}

\maketitle

\begin{abstract}
An onboard target detection, tracking and avoidance system has been developed in this paper, for low-cost UAV flight controllers using AI-Based approaches. The aim of the proposed system is that an ally UAV can either avoid or track an unexpected enemy UAV with a net to protect itself. In this point of view, a simple and robust target detection, tracking and avoidance system is designed.
Two open-source tools were used for the aim: a state-of-the-art object detection technique called SSD and an API for MAVLink compatible systems called MAVSDK. The MAVSDK performs velocity control when a UAV is detected so that the manoeuvre is done simply and efficiently. The proposed system was verified with Software in the loop (SITL) and Hardware in the loop (HITL) simulators. The simplicity of this algorithm makes it innovative, and therefore it should be used in future applications needing robust performances with low-cost hardware such as delivery drone applications.
\end{abstract}

\section{Introduction}

Unmanned aerial vehicles (UAVs) have a growing impact on society. From their first use for war-fighting in 1849, nowadays they are used everywhere: surveillance, delivery, defence, agriculture, and so on. Therefore, their introduction into our daily life arouses safety issues. Today, fully autonomous UAVs are much more efficient than piloted UAVs, but also riskier. One important requirement is the ability to sense and avoid any obstacle in the environment.

Target detection, tracking and avoidance system for UAVs are being developed in different areas: in defence as weapons, in civil as delivery drones, etc. Furthermore, Anti-UAV Defense Systems (AUDS) has been enhanced to counter new threats of UAVs (e.g. Gatwick Airport UAV incident in December 2018). Thus, target tracking is also an essential ability to be developed for UAVs.

BAE Systems sponsored an inter-university UAV Swarm competition to simulate the military-world conditions in a game of offence and defence using UAVs. The mission statement was: ``Create a novel and innovative solution with respect to defending against a swarm of UAVs''. 

In this UAV swarm, two essential requirements had to be respected:

\begin{itemize}
	\item The potential collisions must be detected.
	\item The ally UAVs must be able to either track or avoid enemy UAVs (track the enemy when our agent has a net and avoid the enemy after it has released the net).
\end{itemize}

Before a target is avoided or tracked, it has first to be detected. Many state-of-the-art detection algorithms have been developed in recent years and based on Convolutional Neural Networks (CNNs). There are YOLO~\cite{YOLO}, SSD~\cite{SSD}, R-CNN~\cite{RCNN} and Faster R-CNN~\cite{FasterRCNN} with outstanding performance in both accuracy and frame rates. Hossain et al.~\cite{HOSSAIN} compared onboard embedded GPU systems, off-board GPU ground station, and onboard GPU constraint systems for target detection from a UAV in terms of frame rates and accuracy performance. Due to transmission capacity constraint and the limited onboard computing power, Intel Movidius Neural Compute Stick (NCS) using the Intel Movidius Neural Compute SDK (NCSDK) is introduced to accompany with Raspberry Pi while Mobilenet-SSD (5FPS) outperforms YOLO (1FPS).

When detecting a UAV, a depth distance is primordial to obtain the distance between the UAV and the target. There are multiple active detection sensors, such as radar, LiDAR; however, the cost and deficiency in calibration with the camera lead us to the passive detection via stereo vision. To obtain the depth of the stereo vision, the most effective approach is based on supervised learning in CNNs, which requires an abundant training set for a promising result applied in a single picture. Aguilar et al.~\cite{AGUILAR} proposed a novel unsupervised technique that first calculates disparity in an RGB image via CNNs and then depth by the geometrical relationship with disparity. 

\begin{figure*}[pb]
	\centering
	{\includegraphics[width=0.7\linewidth]{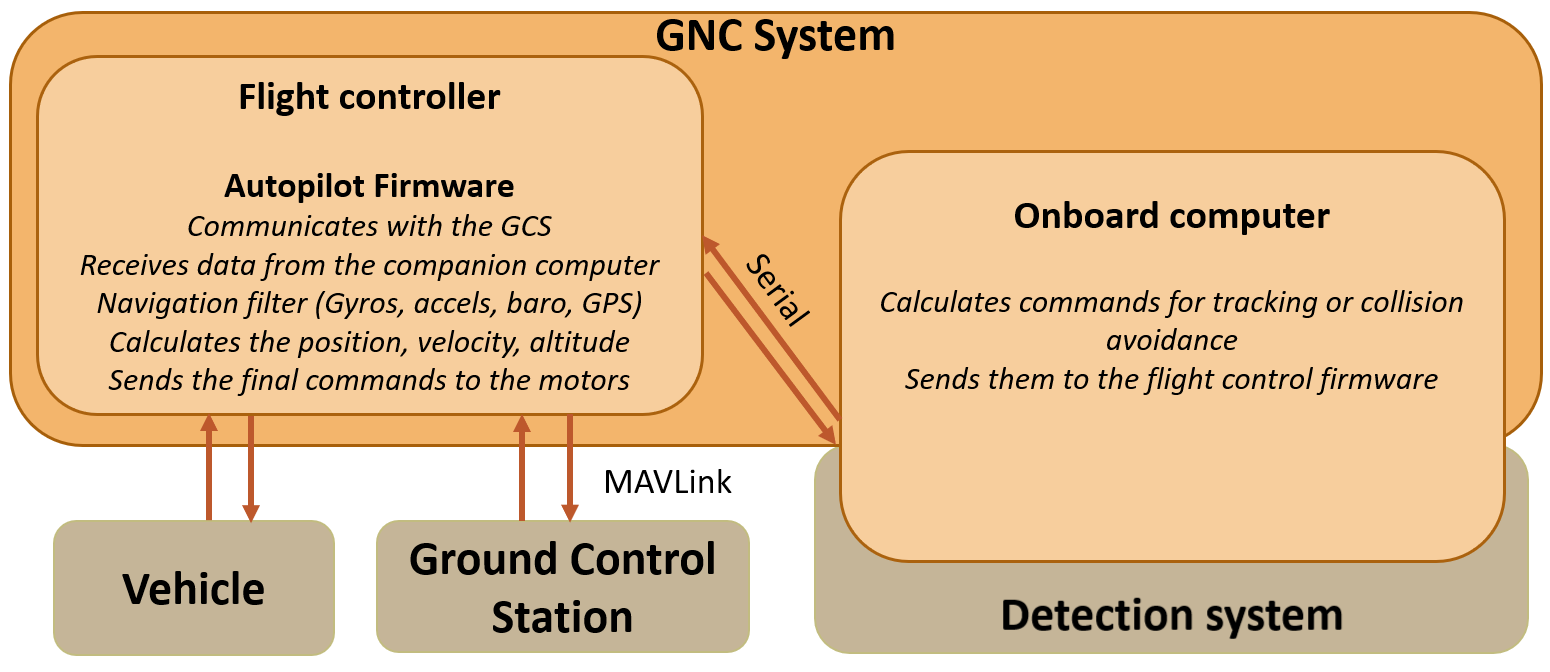}}
	\caption{GNC system architecture}
	\label{architecture}
\end{figure*}

Finally, after a target is detected, it can be avoided or tracked by performing an adapted manoeuvre. In~\cite{SURVEY_AVOID}, a survey on different manoeuvre approaches was done: a geometric approach~\cite{GEO}, an optimised trajectory approach~\cite{OPTI}, a bearing-angle based approach~\cite{BEAR}, a force-field approach~\cite{FORCE}, and so on. The advantage of geometric based approaches compared to other approaches like probabilistic and optimisation based algorithms is that they require less computational power. 

In this paper, our objectives are to obtain a simple but effective approach to the detection, tracking and avoidance problems with low computational power. The paper starts with an overview of the overall architecture, followed by a focus on the detection, tracking and avoidance methodologies. The proposed system is verified with the software in the loop (SITL) and hardware in the loop (HITL). Their performances are assessed in the results and discussions results. Finally, the main outcomes are summarised in the conclusion.

\section{PROPOSED METHODOLOGY}

\subsection{Architecture}

The global architecture for the Guidance, Navigation and Control (GNC) system is represented in Figure~\ref{architecture}. The Ground Control Station (GCS) integrates the graphic user interface (GUI) allowing a user to set the mission parameters and a task control algorithm which is running to create waypoints. The whole ground segment can send the commands to the flight control computer (FCS) via a WiFi network using the MAVLink protocol.

Also, each MAVLink message is bypassed by the FCS and forwarded to the companion computer using a serial connection. The companion computer calculates commands for tracking and collision avoidance using the detection results from the onboard detection system and sends the commands to the FCS. The detection algorithm is running on the companion computer with a vision camera. To access the data from the serial port on the onboard computer side, a MAVLink library with APIs for C++, MAVSDK, has been used.

Hence there are two tracking modes: the waypoints tracking mode using GCS' commands and the target tracking mode using the companion computer's commands. The GNC system determines an appropriate tracking mode according to the onboard target detection results. For example, when detected the target, the GNC system is switching the tracking mode to target tracking, and vice versa.

\subsection{Detection algorithm}
The system requirements of the detection algorithm are essentially based on the following three aspects:

\begin{enumerate}
	\item Real-time ($2$ FPS according to the physical restrictions from CA/tracking algorithms).
	\item Accuracy higher than $70$\% for 50 meters.
	\item Depth distance estimation: Error within one meter for $5$ meters away detection (considering net size).
\end{enumerate}

Considering the requirements of real-time, Mobilenet-SSD brings the best accuracy trade-off within the fastest detectors. Every object that would be likely to be avoided had to be included in the training dataset, and for that, images were mainly taken from the Internet (Google Images, already-made datasets, etc.). For the different sets, the following ratio was followed: training set ($64\%$), validation set ($16\%$) and test set ($20\%$). For example, the most important category is the UAV class and therefore, $2,000$ UAV pictures, labelled by ourselves, were included in the training dataset. The model was trained within TensorFlow in $200,000$ steps, which achieved over 95\% accuracy for the test dataset (different images from the training dataset), with the \textit{mean average precision} metric. Using OpenVino Model Optimizer developed by Intel, TensorFlow model is converted into IR format, which could be interpreted by Neural Compute SDK (NCSDK) and run on the onboard computer.

Since the major challenge of using a camera alone is the fact that no direct depth information is obtained since the angle subtended is the same. However, if we can classify what an object is (e.g. a UAV), and thus a look-up table can be used, and therefore we can use a stadiametric rangefinding method. Every trained class is inside the look-up table and hence, if an object is detected, the rangefinding algorithm will estimate its depth.

\begin{figure}[pt]
	\centering
	{\includegraphics[width=0.6\linewidth]{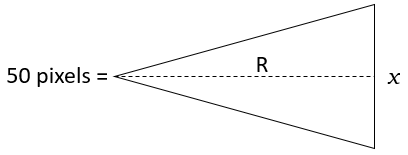}}
	\caption{Stadiametric rangefinding}
	\label{stadiametric}
\end{figure}

In other words, in Figure~\ref{stadiametric}, three parameters are known:
\begin{itemize}
	\item The distance $x$, which is the mean size of both small and big UAVs.
	\item The angle subtended (on the detector, this is occupying for example 50 pixels across).
	\item The instantaneous field of view ($IFOV$) of the detector.
\end{itemize}

Therefore, the subtended angle of the target is $50*IFOV$. Then, basic trigonometry can be used to find the depth of the object. This method requires a good classification of UAVs. In order to obtain the depth estimation function, the linear relationships between the observed size in the camera frame and the actual distance are investigated under the circumstance with different positions in the camera frame. Several linear relationships are obtained from a variety of sizes of UAVs which are averaged to find the middle fitting line. Algorithm \ref{detection_algo} depicts the target detection algorithm which calculates the position of the target.

\begin{figure*}[pb]
	\centering
	{\includegraphics[width=0.7\linewidth]{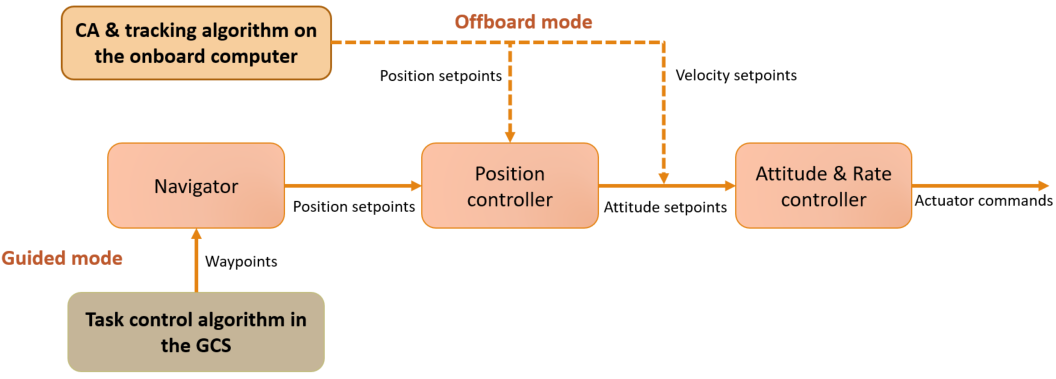}}
	\caption{Guidance architecture}
	\label{guidance_architecture}
\end{figure*}

\begin{algorithm}[pt]
	\caption{Target positioning algorithm using the detection results and depth estimation algorithm}\label{detection_algo}
	\begin{algorithmic}[1]
		\IF{\texttt{a UAV is detected \textbf{AND} the probability is more than a specified threshold}}
		\STATE $W_I \gets \text{Initial width}$
		\STATE $H_I \gets \text{Initial height}$
		\STATE $(x_{min},x_{max},y_{min},y_{max}) \gets \text{Coordinates of the de-}$
		\STATE $\text{tection bounding-box (multiplied by $W_I$ and $H_I$)}$
		\STATE $centre= (x_{min} + \frac{x_{max} - x_{min}}{2}, y_{min} + \frac{y_{max} - y_{min}}{2})$
		\STATE $centre_{modified}= (\frac{2}{W_I}*(centre[0]- \frac{W_I}{2}), \frac{2}{H_I}*(\frac{H_I}{2}- centre[1]))$
		\STATE $size =  \frac{ (x_{max} - x_{min})(y_{max} - y_{min})}{W_I * H_I}$
		\IF{$size < 0.5$}
		\IF{$size \leq 0.04$}
		\STATE $z_D = -120 * size + 7.2$
		\ENDIF
		\IF{$0.04 < size \leq 0.4$}
		\STATE $z_D = -3.42 * size + 2.2$
		\ENDIF
		\IF{$size > 0.4$}
		\STATE $z_D = -3.25 * size + 0.5$
		\ENDIF
		\STATE $x_D = \tan(centre_{modified}[0] * 31.1*\frac{\pi}{180}) * z_D$
		\STATE $y_D = \tan(centre_{modified}[1] * 24.4*\frac{\pi}{180}) * z_D$
		\ENDIF
		\STATE $time.sleep(T_D)$
		\ENDIF
		
	\end{algorithmic}
\end{algorithm}

\subsection{CA/Tracking algorithm}

The requirements of Collision avoidance and Target tracking algorithms are focused on:

\begin{enumerate}
	\item Manoeuvre controllability: a smooth switch between a predefined path (Task control algorithm) and an unexpected target on the trajectory (CA/Tracking algorithm).
	\item Target detection: Tracking when the UAV has a net, and Avoidance when the UAV has no net.
	\item Efficient tracking: To avoid unnecessary manoeuvre if a target is detected in the centre of the camera frame.
\end{enumerate}

Given the first requirement, two ways for UAVs guidance were defined (Figure~\ref{guidance_architecture}):

\begin{enumerate}
	\item Guidance using the Task control algorithm (guided mode): The centralised task control algorithm running on the ground sends inputs to the flight controller by complying with the overall attack and defence strategy.
	\item The tracking \& collision avoidance algorithms running on the onboard companion computer sends direct inputs to the flight control stack using the Offboard mode.
\end{enumerate}

The task allocation algorithm is the default one: the UAV follows a pre-planned mission which depends on its role (attack or defence). The CA/Tracking algorithm, when activated, overrides the commands from the GCS: if the target is detected, the GNC system immediately calculates position or velocity setpoints to either track either avoid, switches the FCS' mode into Offboard mode and sends the setpoints to the position controller (if position setpoint) or before the attitude \& rate controller (if velocity setpoint). 

Hence, the Offboard mode enables to bypass the autopilot for either the position or velocity setpoint, a trade-off (Table~\ref{setpointcontrol}) had to be done considering our application.

\begin{table}[tp]
	\caption{Pros and cons of the Offboard velocity and position control}
	\begin{center}
		\begin{tabular}{ |c|c|c| } 
			\hline
			& Velocity control  & Position control \\
			\hline
			Controllability     & Better & Worse \\ 
			\hline
			Guidance reactivity & Faster & Slower \\ 
			\hline
			Complexity & 2 & 1 \\
			\hline
		\end{tabular}
		\label{setpointcontrol}
	\end{center}
\end{table}

\begin{itemize}
	\item For the bypass of the velocity, the UAV is more controllable and the guidance is faster since the inner part of the controller is modified. However, more parameters need to be selected: the velocity and a shut-off parameter.
	\item For the bypass of the position, the UAV is less controllable and the guidance is slower (outer part of the controller), but it is less complex since only one parameter needs to be controlled and then the autopilot does the rest.
\end{itemize}

\begin{figure}[t]
	\centering
	{\includegraphics[width=0.3\linewidth]{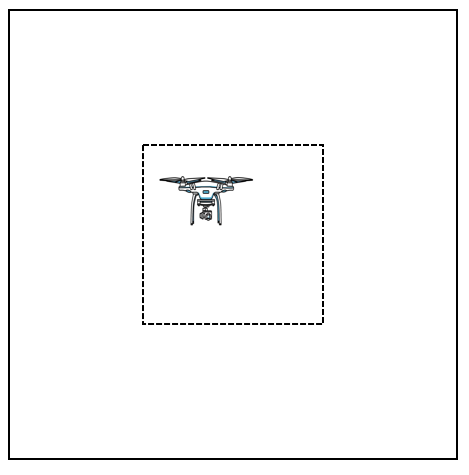}}
	\caption{Threshold of 40\% (dashed box) in the camera frame (solid box)}
	\label{threshold}
\end{figure}

A standard velocity is used when performing the Offboard mode position control. Since robust and fast guidance is needed for sense and avoid purposes, the velocity control is chosen in the Offboard mode in this paper. Furthermore, a minimum security distance margin is needed to respect the third requirement. Thus, a threshold of 40\% was implemented depending on the position of the target in the camera frame, as illustrated in Figure~\ref{threshold}.

Finally, as soon as the detection algorithm stops sending data to the tracking and collision avoidance algorithm, the flight controller firmware automatically switches back into the guided mode. In fact, a time refreshment parameter $t_R$ was introduced in the Task allocation algorithm which continuously searches for Offboard commands when the Offboard mode is activated. Hence, when a target is detected, a velocity setpoint is converted into a MAVLink Offboard command, and if no other target is detected for a time $t=t_R$, the Task allocation algorithm will control the UAV guidance.

A similar timer $T_D$ was also introduced in the detection and CA/tracking algorithms. Indeed, when a target is detected, a specific time is needed for the CA/tracking algorithm to do the manoeuvre. For example, if the detection algorithm gives two consecutive outputs, two consecutive velocity setpoints will be sent to the UAV, and it is physically impossible for the UAV to manoeuvre correctly. Therefore, this parameter was introduced to avoid any manoeuvring instability. The implemented algorithm is described in Algorithm 2.

Hence, when an enemy UAV is detected, the detection algorithm calculates the coordinates in the camera frame and the depth distance estimation and translates them into body-frame coordinates. The latter are then transformed into velocity setpoints in the NED frame (Offboard mode). The simplicity of the algorithm makes the model very efficient even with low-cost hardware.

\begin{algorithm}[pt]
	\caption{Collision avoidance (no net) and Target tracking (with a net and a distance threshold) algorithm}\label{euclid}
	\begin{algorithmic}[1]
		\STATE $S_{x,min},S_{x,max},S_{y,min},S_{y,max} \gets \text{Security threshold}$
		\STATE $(k_1,k_2) \gets \text{Tuned parameters for the velocity}$ 
		\STATE $k_{net} \gets \text{Tuned parameter for the velocity and net}$
		\IF{\texttt{an enemy UAV is detected}}
		\STATE $(x_D,y_D,z_D) \gets \text{Detection output}$
		\STATE $(x_b,y_b,z_b) = (z_D,x_D,y_D)$ \#\textit{Body-frame coordinates}
		\STATE $\phi \gets \text{Heading angle}$
		\STATE $set\_velocity\_ned(0,0,0)$ \#\textit{start Offboard mode}
		\IF{\texttt{our UAV has no net}}
		\STATE $(n,e,d)=(k_1*\sin \big( \phi * \frac{\pi}{180}\big)*x_b,-k_1 * \cos \big(\phi * \frac{\pi}{180} \big)*y_b, k_2*z_b)$ 
		\ENDIF
		\IF{\texttt{our UAV has a net} \textbf{AND} $(x_I<S_{x,min}$ \textbf{OR} $x_I>S_{x,max})$ \textbf{AND} $(y_I<S_{y,min}$ \textbf{OR} $y_I>S_{y,max})$}
		\STATE $(n,e,d)=(-k_1 * \sin \big( \phi* \frac{\pi}{180}\big)*x_b,k_1 * \cos \big(\phi * \frac{\pi}{180} \big)*y_b, k_{net}*z_b)$ 
		\ENDIF
		\IF{\texttt{our UAV has a net} \textbf{AND} $(x_I>S_{x,min}$ \textbf{OR} $x_I<S_{x,max})$}
		\STATE $(n,e,d)=(0 , 0, k_{net}*z_b)$ 
		\ENDIF
		\IF{\texttt{our UAV has a net} \textbf{AND} $(y_I>S_{y,min}$ \textbf{OR} $y_I<S_{y,max})$}
		\STATE $(n,e,d)=(-k_1 * \sin \big( \phi* \frac{\pi}{180}\big)*x_b,k_1 * \cos \big(\phi * \frac{\pi}{180} \big)*y_b, 0)$ 
		\ENDIF
		\STATE $set\_velocity\_ned(n,e,d)$ 
		\ENDIF
	\end{algorithmic}
\end{algorithm}

\subsection{Hardware and software selection}
The GNC system is based on two sub-systems: the flight controller and the companion computer. Given the previous requirements, PX4 was selected for the autopilot firmware. Indeed, it provides a full flight control algorithm and is part of the same project as QGroundControl that was selected as a ground station. Moreover, compared to ArduPilot, which is the other possible solution, PX4 provides more flight modes, which allows the task control algorithm to use unit tasks. Finally, its Offboard mode allows us to send inputs coming from a companion computer directly to the flight control stack, which will be very useful for the tracking and the avoidance.

\begin{figure}[pb]
	\centering
	{\includegraphics[width=0.6\linewidth]{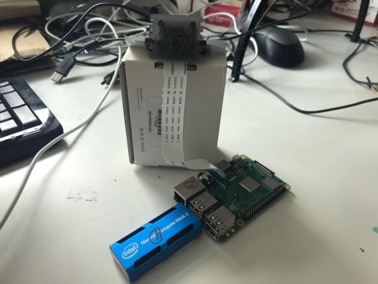}}
	\caption{Hardware selection for the detection system}
	\label{hardware}
\end{figure}

\begin{figure*}[pt]
	\centering
	{\includegraphics[width=1\linewidth]{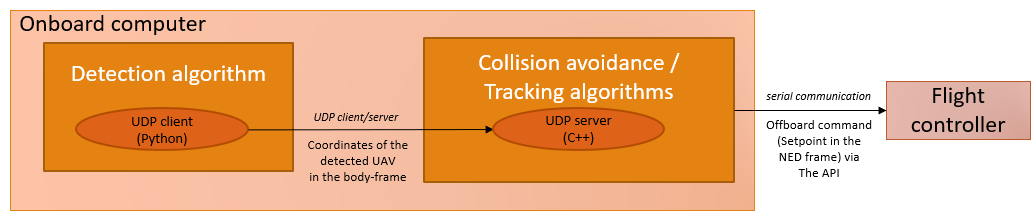}}
	\caption{Architecture of the integrated onboard system}
	\label{udpserver}
\end{figure*}

For the companion computer (seen in Fig \ref{hardware}), the Raspberry Pi 3B+ was selected in accordance with the Vehicle \& Payload operations team and the Detection team, as it can achieve several kinds of operations, like the computer vision, the derived onboard guidance and the payload operation (e.g. net).

For the collision avoidance and tracking algorithms, MAVSDK was used, which is an API that can allow communication between the Pixhawk and the Offboard computer (Raspberry Pi), via a serial communication. It was chosen for different reasons:

\begin{itemize}
	\item Compatibility with the flight controller (Pixhawk), especially with the Offboard mode.
	\item The reliability and information flow between the onboard detection algorithm and the CA/Tracking algorithm.
\end{itemize}

Moreover, since the computational power of the companion computer is poor, the Intel Movidius Neural Network Stick 2 (NCS2) was chosen to work with Raspberry Pi 3B+ as our platform to deploy the Deep Neural Network (DNN). The Raspberry Pi camera module V2 was chosen due to its being lightweight and high compatibility with the RPI.

\subsection{System integration}
The architecture of the integrated onboard system is shown in Figure \ref{udpserver}. Inside the companion computer, UDP sockets were used to link the Detection (UDP client) and the MAVSDK interface (UDP server) since the languages are different (Python and C++). The UDP server can also receive the data coming from the detection program. Using the data from the flight controller and the one of detection, the MAVSDK interface can implement the collision avoidance and tracking algorithm. 

These algorithms first compute the appropriate commands given an input of target position and depth distance estimation within the camera frame, then transformed into coordinates in the body-frame. Then, it forwards a velocity setpoint to the flight controller using the MAVSDK interface to send MAVLink message by the serial connection.

\begin{figure}[pb]
	\centering
	{\includegraphics[width=0.9\linewidth]{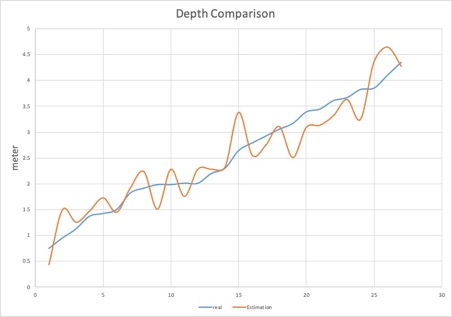}}
	\caption{Depth estimation test results}
	\label{depth com}
\end{figure}

\section{RESULTS AND DISCUSSIONS}

\subsection{Target detection}

According to the system requirements mentioned before, several tests are designed as shown below:

\begin{itemize}
	\item \textit{Detection algorithm test}
	
	The first test was done with the camera module V2 indoor, and the validation results showed that it can run in approximately 18 FPS, which is considerably real-time compared to Faster R-CNN (5FPS), R-FCN (12FPS), YOLOv3-608 (20FPS) and YOLO-tiny (220FPS) running in GPU computers. Besides this, more than 80\% of accuracy prediction is shown stably within 3 meters with a noisy background.
	
	\item \textit{Detection range test}
	
	The detection range is tested through the video knowing the distance between the camera and the target. In the video, UAVs are detected with more than 70\% prediction for a range up to 10-12 meters, where setting 70\% as our boundary of the output.
	
	\item \textit{Depth estimation test}
	
	The estimated depth from the trained model is compared with real distance given by the 3D ground truth. In Figure \ref{depth com}, the orange line is the distance from the ground truth and blue line is the distance from the estimation algorithm. Within 2.5 m, the error is within 0.5 meters, and in the range of 2.5-5.0m, the error is within 1 m, which is acceptable considering the size of the net (1 x 2 m).

\end{itemize}

\subsection{CA/Tracking}
\textit{Testing procedure:} Software and hardware in the loop test were conducted, as well as real UAV tests. For the SITL/HITL tests, the jMAVSim environment was used, which enabled us to simulate a UAV in the Offboard mode, thus different inputs were sent to the Avoidance/Tracking algorithms, then the behaviour of the simulated UAV was observed.

\begin{figure}[!pb]
	\centering
	{\includegraphics[width=1\linewidth]{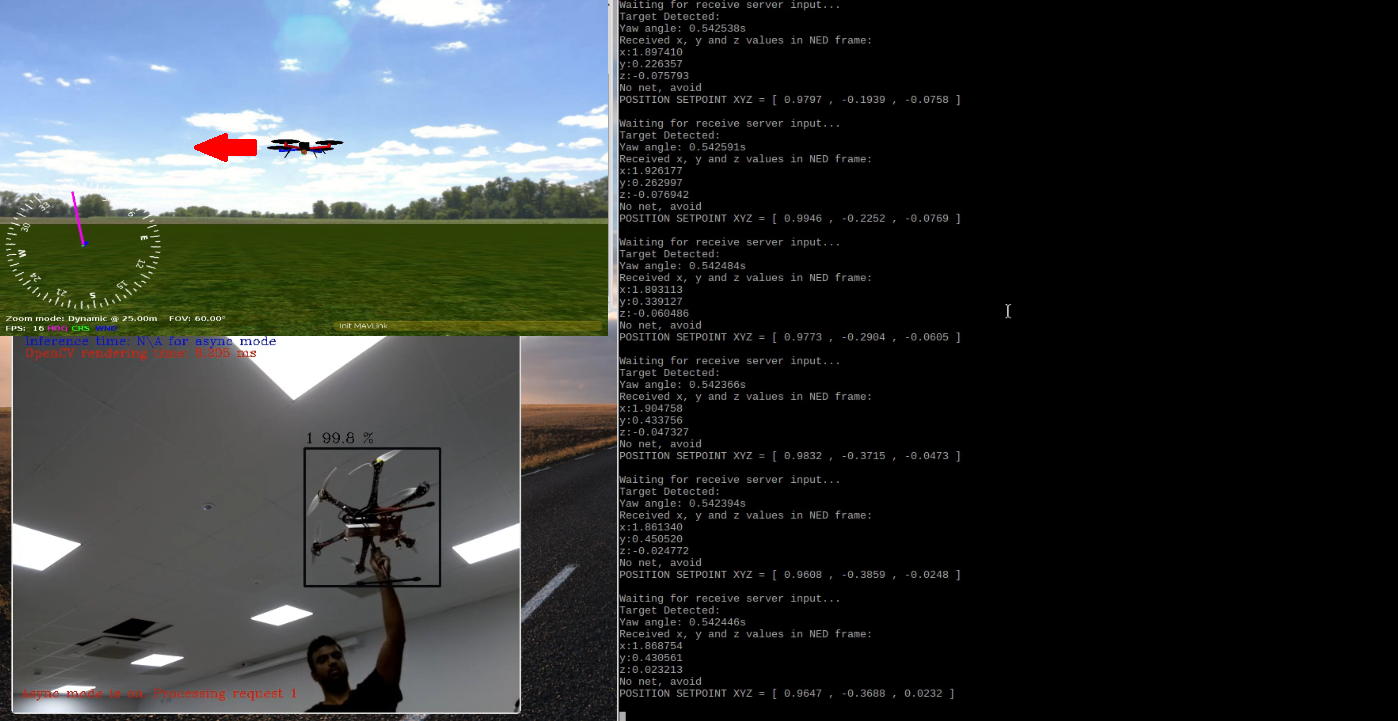}}
	\caption{A capture frame during a real-time simulation of the integrated Detection and Avoidance algorithm using jMAVSim (detected UAV on the right side and therefore the simulated UAV going left)}
	\label{vandv_colavoid_uav_left}
\end{figure}

\begin{figure}[!pt]
	\centering
	{\includegraphics[width=1\linewidth]{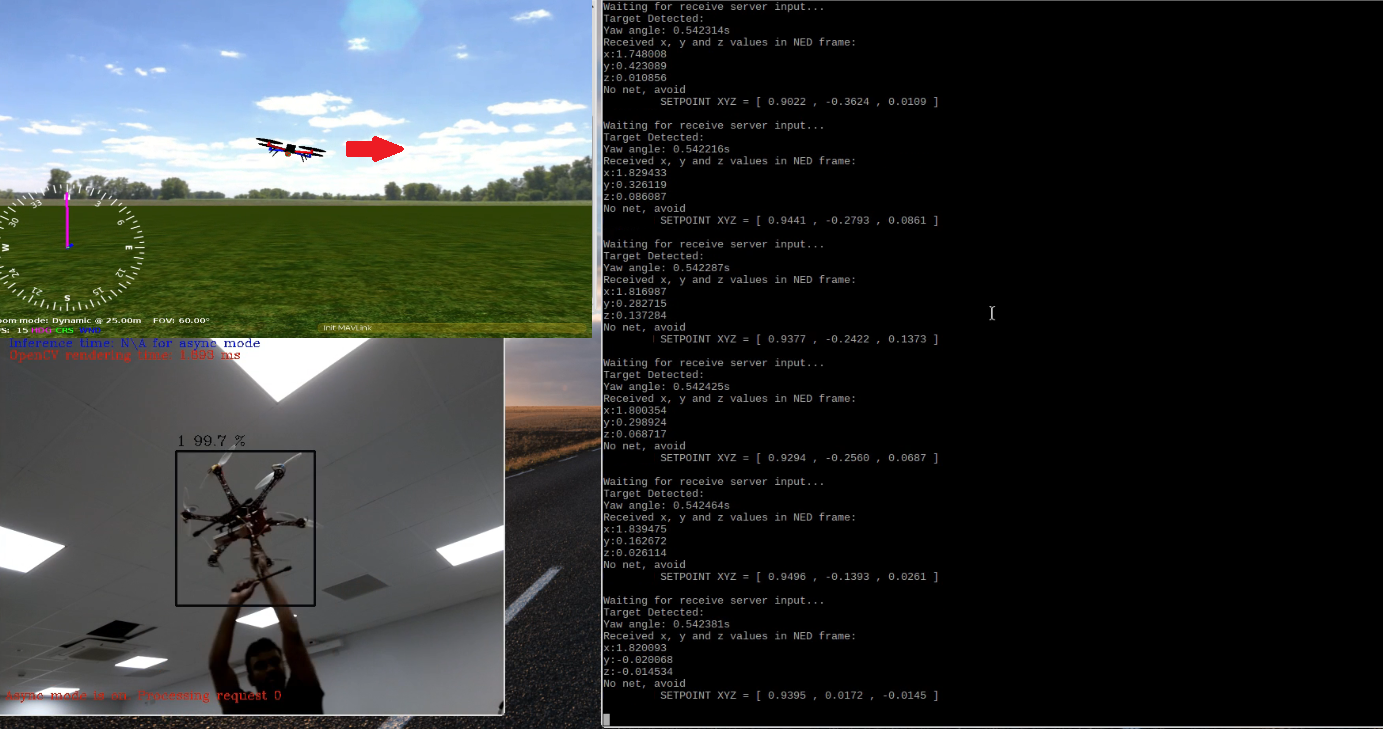}}
	\caption{A capture frame during a real-time simulation of the integrated Detection and Avoidance algorithm using jMAVSim (detected UAV on the left side and therefore the simulated UAV going right)}
	\label{vandv_colavoid_uav_right}
\end{figure}

In Figure~\ref{vandv_colavoid_uav_left} and in Figure~\ref{vandv_colavoid_uav_right}, the detection system sends consecutive outputs from the detected UAV in the camera. These outputs are sent to the Collision avoidance algorithm, they are then translated into velocity setpoints so that the motions of the simulated UAV can be observed for different kind of inputs thanks to jMAVSim. As a result of these tests, the Offboard commands were respected, indeed, the simulated UAVs behaved as expected in the simulation. For instance, when an obstacle was detected in the left area of the camera frame (Figure~\ref{vandv_colavoid_uav_right}), the simulated UAV avoided it by quickly manoeuvring to the right side. 

Moreover, for the tracking algorithm, the motions will be expected to go to the exact opposite, by also taking into account the size of the net. Thus, verification and validation of the tracking algorithm were successfully done in a similar way.

\section{CONCLUSIONS}

This paper proposed a simple target detection, tracking, and avoidance algorithm which is capable of giving good performances with low computational power and consequently running on low-cost hardware. The developed method works well with a low-cost flight controller (Pixhawk) and a low-cost onboard computer (Raspberry Pi). The proposed system was verified with software in the loop test and hardware in the loop test, and the results showed that the UAV could easily avoid the target (or track it when it has a net) with simple and effective manoeuvres. In general, this approach is showed to be a promising tool in considered contexts.

\section*{Acknowledgment}
The authors would like to express their deep gratitude to BAE Systems for their supports on the Swarm Challenge.

\bibliographystyle{unsrt}
\bibliography{library}

\end{document}